\documentclass{article}

\usepackage[final]{neurips_2019}

\usepackage[utf8]{inputenc}
\usepackage[T1]{fontenc}
\usepackage{hyperref}
\usepackage{url}
\usepackage{booktabs}
\usepackage{amsfonts}
\usepackage{nicefrac}
\usepackage{makecell}
\usepackage{amsmath}
\usepackage{microtype}
\usepackage{graphicx}
\usepackage{xcolor}
\usepackage{subfigure}
\usepackage{lipsum}
\usepackage{bbm}
\usepackage{float}
\usepackage{adjustbox}
\usepackage{wrapfig}
\usepackage{caption}
\usepackage{adjustbox}
\newcommand{\ssymbol}[1]{^{\@fnsymbol{#1}}}

\title{
  {SEER-MoE: Sparse Expert Efficiency through Regularization for Mixture-of-Experts} \\
  \vspace{1em}
}

\author{
  Alexandre Muzio\textsuperscript{\textdagger} \qquad 
  Alex Sun \textsuperscript{\textdaggerdbl}\qquad 
  Churan He \textsuperscript{\textdaggerdbl} \\
  Stanford University \enspace \textsuperscript{\textdagger}Google \enspace \textsuperscript{\textdaggerdbl}NVIDIA\\
}

\begin{document}

\maketitle

\begin{abstract}
 % In our work, we aim to investigate and study efficient fine-tuning strategies for sparse Mixture-of-Experts (MoEs) models. Specifically, we propose a two-stage approach where the first one focuses on selecting the best subset of experts per layer and the second one focuses on fine-tuning the gating network to reduce the number of top-k experts queried for each token (thereby reducing FLOPs). We expect that our approach will bring improvements to MoE performance on downstream tasks compared with the vanilla pretrained baselines combined with other model agnostic fine-tuning strategies such as LoRA.
 The advancement of deep learning has led to the emergence of Mixture-of-Experts (MoEs) models, known for their dynamic allocation of computational resources based on input. Despite their promise, MoEs face challenges, particularly in terms of memory requirements. To address this, our work introduces SEER-MoE, a novel two-stage framework for reducing both the memory footprint and compute requirements of pre-trained MoE models. The first stage involves pruning the total number of experts using a heavy-hitters counting guidance, while the second stage employs a regularization-based fine-tuning strategy to recover accuracy loss and reduce the number of activated experts during inference. Our empirical studies demonstrate the effectiveness of our method, resulting in a sparse MoEs model optimized for inference efficiency with minimal accuracy trade-offs.

\end{abstract}

\section{Introduction}

Recent advances in deep learning have propelled the field towards increasingly large and complex models to achieve state-of-the-art performance across a myriad of tasks. Among these, Mixture-of-Experts (MoEs) models have emerged as a promising architecture, distinguished by their ability to dynamically allocate computational resources based on the input ~\citep{fedus2021switch, zoph2022stmoe, jiang2024mixtral, reid2024gemini}.
This paradigm, characterized by a sparse gating mechanism that routes inputs to a subset of specialized expert networks, allows for the scalable expansion of model parameters while maintaining a relatively constant computational footprint per input token.

However, the path forward for MoEs involves addressing significant challenges, particularly regarding their substantial memory requirements. The advent of very large MoE models such as Grok-1 \cite{xAI2024Grok}, with 314B parameters distributed across 8 experts, underscores the urgency of addressing this issue. While the sparse nature of MoEs promises enhanced efficiency and scalability, the sheer size of the models introduces a new set of complexities, particularly in terms of memory footprint.

In light of these challenges, our work aims to investigate and develop pruning and fine-tuning strategies that dynamically adjust the quantity and allocation of experts in an MoE-based language model. 

The main contributions of our paper is an in-depth study of Parameter Count / FLOPs for MoE models and the SEER-MoE method, a novel 2-stage approach that takes a step in the direction of reducing the memory footprint / compute requirements for pretrained MoE models. Our first stage proposes to prune the total number of experts in the model with a novel \textit{heavy-hitters counting} guidance to reduce the memory footprint for loading the entire MoE model. Our second stage proposes an effecitve regularization-based finetuning strategy to recover the accuracy loss from previous pruning while simultanously reducing the number of activated experts during inference. The combination of both stages yields a sparse MoEs model with cheaper memory requirements while being optimized for inference efficiency, at the compensation of minimized accuracy drops. 

We perform extensive empirical studies with the popular Mixtral 8x7b~\cite{jiang2024mixtral} MoEs model on both SST5~\cite{socher-etal-2013-recursive} and MMLU~\cite{hendrycks2020measuring} to validate the effectiveness of our method, including an in-depth ablation study to understand different design choices for each of our stage.

\section{Related Work}

\textbf{Mixture-of-Experts (MoE)} dates back at least since work from \citet{JacobsMoE1991}, which introduce a new model architecture composed of many separate networks and each one handles a subset of the complete set of training cases. Each expert specializes in a different region of the input space. \citet{david2014deepmoe} extended the Mixture-of-Experts to a stacked model with multiple layers of gating and experts, and exponentially increases the number of effective experts through layers of combination. As the rapid advancement of LLM, MoE ~\citep{jiang2024mixtral, reid2024gemini, xAI2024Grok} gained increased popularity for it's scalability, efficiency and STOA evaluation result from various benchmarks. Most MoE architectures \cite{fedus2021switch, zoph2022stmoe, jiang2024mixtral} includes a specific gating network that learns the optimal routing from input tokens to experts. However, the weights of the gating network stay fixed, regardless of what task is being solved. To the best of our knowledge, there is no such method that explores Top-K routing adaptation.

\textbf{Sparsification} aims to remove certain parts of the network. In Mixture-of-Experts models, only a few experts (top-K) chosen by the router will be activated in each layer to generate output, therefore removing unused experts can linearly reduce the model size without loss in performance.
\citep{lu2024efficientpruning} introduced a heuristic search method to prune the number of experts in post-training. The method is based on the enumeration of expert combinations and choosing the target eliminating experts based on the lowest reconstruction loss. They verified their approach's effectiveness on Mixtral 8x7b \citep{jiang2024mixtral}. However, the method has high time complexity and isn't applicable to models with large expert counts nor cross-layer pruning
\citep{zhang2023h2o} proposed an eviction algorithm targeting on KV cache based on "Heavy Hitters" which holds a significant role for model performance. We hypothesize that a similar strategy in expert activation counting can be applied to MoE pruning.

\section{Methodology}

The computation and resources used by large MoEs models (FLOPs) mainly come from two factors:

\textbf{The total number of available experts at each MoE layer}, denoted as $M_l$ for layer $l$, and determines how much VRAM we need to completely load the model on the GPUs/TPUs.

\textbf{The top-K number of experts activated}, denoted as $k_l$ for layer $l$,  and controls how much computation is used per token. 

% In this work, we start from a pretrained MoE language model (Mixtral 8x7b) and we propose a two-stage approach which attempts to reduce both the number of parameters of the model (via MoE-specifc sparsification) and the number of experts being activated.

In terms of notation, suppose we have a fine-tuning dataset $\mathcal{D}$ which consists of sample $(x_i, y_i)$ pairs, the gating network at layer $l$ is $g_l$, and the $j$th expert at the $l$th layer is denoted as $e_l^j$. Moreover, suppose the features collected at layer $l$ is $f_l(x_i)$.
% In terms of notation, suppose that we have a fine-tuning dataset $\mathcal{D}$ which consists of sample $(x_i, y_i)$ pairs, the gating network at layer $l$ is $g_l$, and the $j$th expert at the $l$th layer is denoted as $e_l^j$. Moreover, suppose the features collected at layer $l$ is $f_l(x_i)$.

\begin{figure}[t]
    \centering
    \includegraphics[width=.8\linewidth]{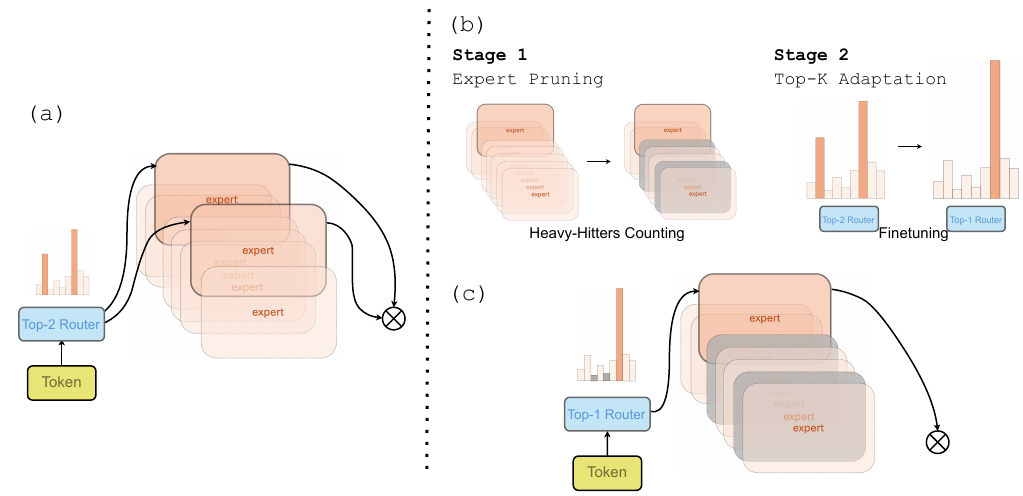}
    \caption{SEER-MoE visualized in a two-stage process. (a) The initial model with all experts and top-2 router. (b) Stage 1 involves expert pruning based on heavy-hitters counting to identify and retain the most critical experts; Stage 2 includes top-K adaptation through fine-tuning to optimize the number of active experts, culminating in a model that balances efficiency and performance. (c) SEER-MoE with expert pruning and top-K adaptation.}
    \label{fig:main_method}
    \vspace{-15pt}
\end{figure}

\subsection{Parameter and Compute Scaling of MoE Transformers}
\label{ssec:flops_analysis}
Following the work of \citet{ScalingLawsOpenAI2020}, we extend the parameterization of the Transformer architecture to that of the sparse MoE architecture (more details in Appendix \ref{app:param_flops_analysis}). We utilize the Transformer hyperparameters together with $n_{experts}$ (number of experts per layer) and $n_{topk}$ (number of experts activated per token). Using $N$ to denote the number of non-embedding parameters and considering $d_{attn} = d_{ff} / 4 = d_{model}$

\begin{equation}
    N \approx 4d^2_{model}n_{layer} (1 + 2 n_{experts})
    \label{eq:n_params_moe}
\end{equation}

And the FLOPs for the forward pass:
\begin{equation}
    C_{fwd} \approx 8d^2_{model}n_{layer} (1 + 2 n_{topk}) + 2n_{layer}n_{ctx}d_{model}
\end{equation}

It is notable that for the MoE blocks of the network, the number of parameters from the MoE Feedforward layers increase in proportion to $n_{experts}$ while the FLOPs per Token to $n_{topk}$. This motivates us to think whether we can target reducing both the compute FLOPs of the model and the memory separately.

% Motivation
Specifically for the Mixtral 8x7B \cite{jiang2024mixtral} model, which utilizes $2$ experts per token, Expert blocks computations account for about $55\%$ of the total FLOPS. Additionally, for a model with the same architecture but with only a single expert being activated, FLOPs reduces by $27\%$. This motivates us to explore whether we are able to adapt existing models to use less compute.

\subsection{Expert Sparsification with Heavy-hitters Counting}
\label{subsec:heavy_hitter}
% Calculate FLOPs for different sparsification ratios
% Memory reduction
Considering the large memory requirement of MoE models described and considering that the bulk of the parameters belong to the Expert layers, we propose reducing the total number of experts $M_l$ at each layer $l$. We do so, by reducing the number of expert models. We propose to carry out this investigation with a data-driven strategy.

For a MoEs model, an expert $e_l^j$ is activated for a token if its corresponding router logit $g_l(f_l(x_i))^j$ is ranked in the top-K after softmax. Specifically, we denote this selection process with a function $\gamma(.)$:
\begin{align}
    \gamma(e_l^j, x_i) &= 1 \text{ if } j \in \text{ArgTopK}(\text{softmax}(g_l(f_l(x_i))), K),\\
    \gamma(e_l^j, x_i) &= 0 \text{ else }
\end{align}

For each expert $e_l^j$, we define the \textit{activation counts} $a_l^j$ as the total number of times it gets activated. Formally, this could be expressed as follows:
\begin{align}
    a_l^j = \sum_{(x_i, y_i) \in \mathcal{D}} \mathbbm{1}[\gamma(e_l^j, x_i) = 1]
    \label{eqn:heavy_hitter}
\end{align}
Equivalently, Eqn.\ref{eqn:heavy_hitter} is performing \textbf{\textit{a Monte-Carlo estimate of the marginal probability $P(e_l^j\text{ gets activated})$}} using the dataset $\mathcal{D}$, which provides theoretical motivation for our adopted technique here.

\textbf{Soft Counting} We also propose another variant with a softer and more relaxed version of heavy-hitters counting. Intuitively, for certain tokens, if expert $e_l^j$ is activated, we don't know whether it wins over other experts by a slight margin or gets activated with high confidence. Since the binary activation count can not capture this exact magnitude of confidence in activating certain experts, we propose to directly leverage the softmax probabilities as soft counts. Formally, this can be defined as:
\begin{align}
    a_l^j = \sum_{(x_i, y_i) \in \mathcal{D}} \text{softmax}(g_l(f_l(x_i)))^j
    \label{eqn:heavy_hitter}
\end{align}

\textbf{Layer Expert Pruning}
Now with the statistics of the heavy-hitters counts, we could leverage them as powerful guidance to remove experts that are unlikely to be activated for data from $P_{\mathcal{D}}$. Suppose we want to keep a total of $\hat{M}_l$ experts per layer, the kept experts at layer $l$ are denoted as:
\begin{align}
    \text{ArgTopK}(\bigcup_{j \in [1, M_l]} \{a_l^j\}, \hat{M_l})
\end{align}

We repeat this for every layer in the MoEs model to get an entire mask.

\textbf{Global Expert Pruning}
Since the counts have uniform magnitude range across all layers, we can also carry out a global sorting and pruning to remove the experts with the least probability of getting activated. Suppose we only want to keep a total of $\hat{M}$ experts in the network. The kept ones are denoted as:
\begin{align}
    \text{ArgTopK}(\bigcup_{l \in [1, L]}\bigcup_{j \in [1, M_l]} \{a_l^j\}, \hat{M})
\end{align}
Compared with the above Layer Expert Pruning option, this will provide a nonuniform expert sparsity pattern across layers but could potentially be of higher-quality.

To recap, for pruning experts from the MoEs model to reduce storage burden and memory footprints, we propose to perform pruning based on heavy-hitters counting. This counting could be either \textbf{\textit{actual activation counts}} or a \textbf{\textit{soft counting with softmax probabilities}}. The actual removal of experts could be conducted either \textbf{\textit{layer-wise}} or \textbf{\textit{globally}}. This gives us a total of four configurations with combinations of pairs of counting and removal strategies, and we are going to provide detailed ablation results in the Experiment section.

\subsection{Enhancing Expert Efficiency: Advanced Finetuning Strategies}
\label{sec:fine-tuning}

With the goal of reducing the number of experts activated for each token during inference, while still maintaining competitive performance,
we propose different fine-tuning procedures for MoE model.

\subsubsection{Top-K adaptation}

Starting from a pretrained model trained with top-k > $1$, we posit that by fine-tuning the model on a downstream task while reducing $k$ during training is a feasible and simple strategy to adapt the model to utilize. We focus on fine-tuning since we are interested in utilizing existing pretrained more efficiently.

Given that we are trying to target the best open-source available MoE model, which is the Mixtral 8x7b 
% \footnote{On March 17th, 2024, on the final deadline of this report, Grok-1 model came out which is the largest open-source MoE model with 314B parameters. Interestingly, this model has a very similar architecture to Mixtral 8x7b model, with 8 experts and top-2 routing which makes this work also applicable for that model.}
standard fine-tuning was not feasible given the amount of memory required (even using 8xA100 80GB), therefore we opted for explore QLoRA \cite{dettmers2023qlora} fine-tuning on the self-attention blocks to reduce the memory footprint of the optimization. 

In this work, we propose Top-K reduction procedures with simplicity in mind: Static top-k with $k < K$ and Annealing top-k from $K \rightarrow k$ with $k < K$.
We also explore additional methods such as QLoRA fine-tuning targeting only the gating network.

\subsubsection{Entropy-based gating regularization}

Entropy, in the context of information theory, is a measure of the unpredictability or randomness of a distribution. 
For a categorical probability distribution (which is the case of the expert gating network), entropy is defined as $H(X) = -\sum_{i=1}^{n} p(x_i) \log p(x_i)$. 
We posit that a gating network with a more peaky distribution, meaning lower entropy, relies more heavily on a single expert. 
Therefore, by minimizing the entropy of the gating network's distribution, we encourage the model to make more decisive selection of experts while reducing the computational overhead associated with activating multiple experts.

The final loss we propose is $loss = \mathcal{L}_\text{cross entropy} + \lambda \mathcal{L}_\text{entropy}$, and explicitly:

\begin{equation}
\label{eq:fine-tuning_loss}
\text{loss} = -\sum_{t=1}^{T} \sum_{i} y_{t,i} \log(p_{t,i}) + \lambda \left(-\sum_{t=1}^{T} \sum_{j} p_{t,j} \log(p_{t,j})\right)
\end{equation}

Here $\lambda$ is a hyperparameter, and the first term denotes the standard cross-entropy loss, and the second term denotes the entropy loss.
The top-2 gating mechanism inherently provides a form of redundancy, which can be beneficial for robustness and handling uncertainty.
Moving to a more peaky, top-1 distribution could reduce this redundancy, potentially making the model more sensitive to errors in the expert selection.

\subsection{SEER-MoE: Sparse Expert Efficiency through Regularization for Mixture-of-Experts}

We introduce SEER-MoE (Sparse Expert Efficiency through Regularization for Mixture-of-Experts), a two-stage approach designed to enhance the computational efficiency of pslightlyed MoE models, specifically targeting the Mixtral 8x7b (but is also applicable to other MoE models) such as Grok-1:

\textbf{Stage 1: MoE-specific sparsification} decreases the total number of available experts ($M_l$) in each MoE layer via effectively pruning the less significant ones.

\textbf{Stage 2: Top-K adaptation} i regularization techniques during fine-tuning, encouraging the model to rely on fewer experts without compromising the quality of the learned representations while better utilizing the information from the experts that have been pruned in Stage 1.

Finally, considering the total Model Parameter/FLOPs analysis from Section \ref{ssec:flops_analysis}, by reducing the number of experts by 25\% and using a single activated expert, we reduce the model parameters by $\approx 25\%$ and FLOPs by $\approx 27\% $ while only slightly degrading model performance.

\section{Experiments}
This section outlines the experimental setup, including model configurations, training details, and evaluation metrics used to assess the performance of our proposed SEER-MoE method.

\subsection{Experimental details}

We utilized the Mixtral 8x7b model for all our experiments, chosen for its superior performance over OpenMoE in our initial tests (for more quality results, check Appendix \ref{app:openmoe}). Our fine-tuning employed QLoRA on 8 NVIDIA A100 GPUs (80GB), running for 1000 steps (unless specified otherwise) using Adam optimizer with a weight decay of 0.01. 

\textbf{Distributed training (Technical challenges with sharding)}. 
Despite facing challenges with distributed training and sharding, we managed to optimize our setup for efficient training.
We utilized QLoRA fine-tuning to minimize memory usage, crucial for our hardware constraints.
We also explored PyTorch's Fully Sharded Data Parallelism (FSDP) \cite{zhao2023pytorch} as a viable option but given that QLoRa was showing good results with focused on QLoRA.
We implemented the full distributed training loop using the \citet{accelerate} library.

\subsection{Data}
We chose MMLU \citep{hendrycks2020measuring} for multitask language understanding with 57 subjects and Stanford Sentiment Treebank-5 (SST5) dataset \citep{socher-etal-2013-recursive} for sentiment classification.

MMLU dataset contains questions, choices, and correct labels. The question and choice pairs are formulated to prompt the following format in Appendix \ref{app:prompts}. The answer will be extracted from model output and compared against the correct label to determine output accuracy, which is used to evaluate various expert sparsification strategies (counting, pruning, subject-specific masking) quantitatively.

SST5 dataset contains texts and sentiment labels (very negative, negative, neutral, positive, very positive). The text is applied to the instruction template from Appendix \ref{app:prompts} to form the prompt to the model. Accuracy will be computed by comparing the extracted label from the model's text output with the ground-truth label. We use the the SST5 training set ($\approx$ 8.5k examples) for fine-tuning model on different configurations (expert sparsity, top-K, fine-tuning precedure) and evaluate on the validation set ($\approx$ 1.1k examples).

\subsection{Evaluation Method}
The evaluation metrics we used are result accuracy and reduced computation (FLOPs) and memory footprint.

\textbf{Accuracy} is measured by the percentage of answers extracted from model output matches with the validation dataset's labels. For MMLU task, we calculate the accuracy from number of correct answer among all questions and total number of 1531 questions in validation set. For SST task, we calculate the accuracy from number of correctly assigned sentiment with the total number of 1101 texts in validation set.

\textbf{FLOPs and Memory Reduction} is calculated based on model sparsification and ablation configuration. The FLOPs per token of the pruned model are calcuated based on Table \ref{table:transformer-flops}. With expert removal, the model would contain less experts thus less parameters. The memory reduction can be calculated by number of expert reduction at each layer.

\section{Results and Analysis}

\subsection{Expert Sparsification with Heavy-hitters Counting}
We start by evaluating the effectiveness of our proposed pruning strategy to reduce the storage and memory footprint of MoE models with Heavy-hitters Counting.
% We are going to first discuss baselines we compare to validate the method then demonstrate the comparison of our \textbf{top-performing variant} with them. Recall we propose several configurations with counting strategies including \textbf{actual activation counts} and \textbf{soft counting} and expert removal strategies of \textbf{layer pruning} and \textbf{global pruning}. We are going to demonstrate ablation results to compare their effectiveness. To conclude this section, we show visualization of our collected counting statistics on MMLU with Mixtral in a heatmap to understand what are the activation as well as sparsity patterns of experts.

\textbf{Baselines}
We have 3 baselines to compare against: Dense baseline without expert, randomly pruning the experts and the state-of-the-art expert pruning stategy proposed by ~\citet{lu2024efficientpruning}.

For clarity, we report the results with the dense model numbers serving as the reference to indicate how much pruning affects the model. In terms of the baseline statistics of the dense model, from our evaluation on MMLU, we get a Mean Accuracy of $60.55\%$ over all subjects and a Memory Usage of $86$GB with $bf16$. Notice that our dense accuracy of Mixtral on MMLU does not exactly match the numbers reported in ~\citet{lu2024efficientpruning} due to potential hardsware mismatch and prompt difference. However, we are comparing with the accuracy drop from the respective dense baselines which ensures a fair comparison on the same scale.

All reported results are zero-shot without fine-tuning.

% Dense 86GB x1
% 25% 65GB 0.76
% 50% 46,879GB 0.55

% speed, 25%: 1.20, 50%: 1.27
\textbf{Comparison}
We demonstrate the comparison of our top-confirming configuration with the baselines in Table~\ref{tab:baselines}. As observed, at either $25\%$ or $50\%$ expert sparsity, we significantly surpass all the baselines by a clear margin, achieving much smaller accuracy drop. For example, compared with the latest SOTA approach~\cite{lu2024efficientpruning}, at the same $25\%$ expert sparsity, we only lose $\mathbf{3.85}$ accuracy points, almost halve the accuracy drop of $6.40$ of ~\cite{lu2024efficientpruning}.

Moreover, we could observe that with the expert sparsity introduced into the model, we successfully reduce the memory footprint for loading Mixtral by $24\%$ with $25\%$ expert sparsity and $45\%$ with $50\%$ expert sparsity. While it is not possible to fully load the entire dense Mixtral $8x7b$ model on a single $80 GB$ $A100$ GPU(Dense: $86 GB$), our proposed apporach provides a viable solution at the compensation of minimized accuracy loss by pruning unimportant experts from the model guided by Heavy Hitters Counting. We are also going to show later that this accuracy drop could be further minimized with Task-Specific Finetuning.

\begin{table}[t!]
    \centering
    \begin{adjustbox}{width=\linewidth}
    \begin{tabular}{c|c c c c}
    \toprule
      Method  & Total Expert Sparsity $\uparrow$& Accuracy Drop from Dense $\downarrow$ &Memory Usage$\downarrow$ & Speedup$\uparrow$\\
    \midrule
    \midrule
    Dense & $0$ & $0$ & $\times 1$ & $\times 1$\\
    \midrule
    Random & $25\%$ & $6.17$ &$\mathbf{\times 0.76}$ & $\mathbf{\times 1.20}$ \\
    ~\cite{lu2024efficientpruning} & $25\%$ & $6.40$ & $\mathbf{\times 0.76}$ & $\mathbf{\times 1.20}$\\
    \textbf{Ours} & $25\%$ & $\mathbf{3.85}$ &$\mathbf{\times 0.76}$ & $\mathbf{\times 1.20}$\\
    \midrule
    Random & $50\%$ & $15.19$  &$\mathbf{\times 0.55}$ & $\mathbf{\times 1.27}$\\
    ~\cite{lu2024efficientpruning} & $50\%$ & $16.12$  & $\mathbf{\times 0.55}$ & $\mathbf{\times 1.27}$\\
    \textbf{Ours} & $50\%$ & $\mathbf{13.78}$ &$\mathbf{\times 0.55}$ & $\mathbf{\times 1.27}$\\
    \bottomrule
    \end{tabular}
    \end{adjustbox}
    \caption{Comparison with baseline approaches. Ours achieves the minimized accuracy drop from the dense baseline at all expert sparsity levels. Notably, we beat the state-of-the-art ~\cite{lu2024efficientpruning} with a clear margin.}
    \label{tab:baselines}
    \vspace{-15pt}
\end{table}

\begin{table}[t!]
    \centering
    \begin{adjustbox}{width=\linewidth}
    \begin{tabular}{cccc| c}
    \toprule
      Method & Counting Strategy & Pruning Strategy  & Subject-Specific Pruning & Accuracy Drop from Dense $\downarrow$\\
    \midrule
    \midrule
    Dense & n/a & n/a & n/a & $0$ \\
    \midrule
    Ours & Activation & Layer & Yes & $16.59$\\
    Ours & Activation & Global & Yes & $14.63$\\
    Ours & Soft & Layer & Yes & $15.54$\\
    Ours & Soft & Global & Yes & $7.90$\\
    Ours & Soft & Layer & No & $12.80$\\
    Ours & Soft & Global & No & $\mathbf{3.85}$\\  
    \bottomrule
    \end{tabular}
    \end{adjustbox}
    \caption{Ablation results of ours. Using soft counting with global pruning and no subject-specific mask yields the best result.}
    \label{tab:ablation}
    \vspace{-25pt}
\end{table}

\textbf{Ablation}
Here, we will compare different heavy hitters counting option and expert removal strategies. Recall that we could use either \textbf{actual activation counts} or \textbf{soft counting} to gather count data and conduct either \textbf{layer pruning} or \textbf{global pruning} to actually remove the experts. Moreover, for MMLU containing $57$ total subjects, we also study whether using a \textbf{Subject-Specific Pruning} strategy benefits the performance. Concretely, with Subject-Specific Pruning, we are going to gather counting statistics and perform pruning for each subject independently, which creates a unique expert mask for solving each subject. Without this option, we gather counting numbers from samples of all subjects and adopt the same expert mask to solve all subjects. We demonstrate comprehensive ablation results in the following Table~\ref{tab:ablation}.

From the table, we could make the following three observations:

\textbf{Observation \#1:} Global pruning works better than layer pruning. Given our count statistics are uniform in magnitude across all layers, global pruning offers more flexible solutions without the layer constraint. Expectedly, we see the results are better with it.

\textbf{Observation \#2:} Using soft counting works better than actual activation counts. Recall that soft counting, by directly accumulating the softmax probability for each expert, gives us a sense of the confidence in selecting each expert to better cope with the scenarios when certain expert barely wins over others. This is validated by the superiorty in results shown in the table.

\begin{wrapfigure}{r}{0.45\textwidth}
% \begin{figure}[t!]
    \centering
    \includegraphics[width=\linewidth]{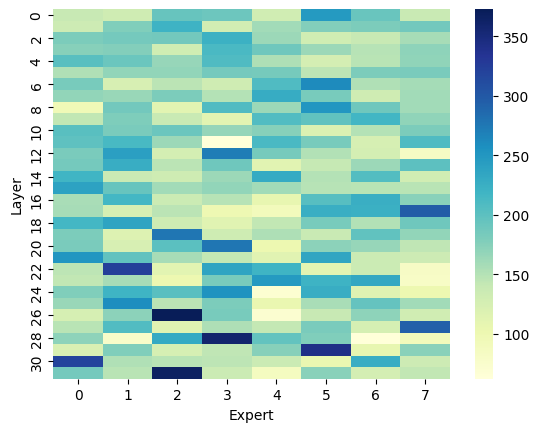}
    \caption{Heavy Hitters Counting Heatmap with Mixtral 8x7b on MMLU.}
    \label{fig:heatmap}
% \end{figure}
\end{wrapfigure}

\textbf{Observation \#3:} Subject-specific pruning does not help to gain better performance on MMLU. Although we expect the subject-specific pruning could offer improvements by varying the expert mask adaptively based on the subject, the results suggest otherwise. There could be a theoretical support for this phenomenon. Recall that as discussed in Sec.~\ref{subsec:heavy_hitter}, heavy-hitters counting is equivalently performing a maginal probability Monte-Carlo estimation. Without subject-specific pruning, more samples are used to build this estimation, which could make it more solid.

\textbf{Visualization and Analysis}
Here, we provide a heatmap visualization of the collected heavy-hitter statistics of Mixtral on MMLU in Fig.~\ref{fig:heatmap}. As observed in the figure, we could observe that some experts are heavily activated and leveraged during inference for example Expert $\#2$ from Layer $26$ and $30$; whereas some experts are barely activated for example Expert $\#7$ from Layer $22$ and $23$. From this discrepancy of acitvation patterns, we could see why heavy-hitters counting could serve as an effective guidance for pruning experts.

% 1. Sparsification

% Ablation table

% "Global or soft counting improves over standard heavy hitters"

% Improve over the Sparsification paper (SOTA?)

% Visualization of Hot Map - to demonstrate intuition

\subsection{Finetuning}

We assessed the efficacy of various strategies for reducing the number of activated experts within the Mixtral 8x7b model by fine-tuning on a sentiment classification task, SST5. The model was finetuned using the training set and evaluated on the evaluation set via text generation employing greedy decoding. For details regarding the prompts we used, refer to Appendix \ref{app:prompts}.

We established a baseline by fine-tuning with top-2 routing to gauge the performance delta attributed to the reduced number of activated experts. This was further compared with a baseline in-context learning approach. We also evaluate the different procedures proposed in Section \ref{sec:fine-tuning}.
Table \ref{table:ft_simple} summarizes our findings.

Our results reveal a performance gap when using a single expert as opposed to two, with zero-shot accuracy dropping by 8.2 percentage points. However, through QLoRA fine-tuning (FT), this gap is mostly bridged, yielding a similar accuracy of 50.8\%. Notably, employing two experts still holds a marginal advantage, as indicated by the 53.6\% accuracy rate post fine-tuning. Nevertheless, when allowed two experts, the QLoRA approach demonstrated a slight improvement over the zero-shot baseline.

\begin{wrapfigure}{r}{0.5\textwidth}
\centering
\adjustbox{width=.5\textwidth}
{
\begin{tabular}{ccc}
\toprule
Method & Top-K & SST5 acc $\uparrow$  \\
\midrule
\midrule
Zero-shot                & 1    &  42.6\% \\ 
Zero-shot                & 2    &  50.8\% \\
\midrule
QLoRA FT             & 1    &  \textbf{50.8\%} \\ 
QLoRA FT             & 2    &  \textbf{53.6\%} \\
\midrule
QLoRA on router    & 1    & diverges \\
QLoRA on router    & 2    &  51.4\% \\
\midrule
FT + Entropy loss, $\lambda=1$    & 1    & 45.5  \\
FT + Entropy loss, $\lambda=0.1$    & 1    & 50.7\% \\
+ Annealing Top-K & 1 & 48.6\%    \\
\makecell{+ Annealing Top-K \\(more steps)}    & 1    &  51.4\% \\
\makecell{+ Annealing Top-K \\+ Entropy Loss}    & 1    &  \textbf{51.8}\% \\
\bottomrule
\end{tabular}
}
\captionof{table}{Comparison of fine-tuning strategies and their impact on SST5 accuracy.}
\label{table:ft_simple}
\vspace{-15pt}
\end{wrapfigure}

These findings suggest that fine-tuning can recover some of the losses attributed to reducing the number of activated experts.
Finally, the approach that works the best combines Annealing with Entropy minimization approach, performing better than naive Top-1 finetuning and only 1.8\% less than Top-2 finetuning.

% To also evaluate the different procedures proposed in Section \ref{sec:fine-tuning}, we do so by also evaluating on SST5.
% Table \ref{table:ft_sst5_ablation} demonstrates our approaches and the first and second rows is the same baseline with fine-tuning.

% \begin{figure}[t!]
%   \begin{minipage}[b]{0.49\textwidth}
%     \centering
%     \includegraphics[width=.85\linewidth]{figs/Ablation_plots.pdf}
%     \vspace{-10pt}
%     \captionof{figure}{\textbf{Ablation study results on ImageNet with ResNet50}. We show results of each improvement acting individually. Top-right is better.}
%     \label{fig:ablation}
%   \end{minipage}
%   \hfill
%   \begin{minipage}[b]{0.49\textwidth}
%     \centering
        
%       \label{table:ablation}
%     \end{minipage}
%     \vspace{-20pt}
% \end{figure}

% Combined with previous table
% \begin{table}[ht]
% \centering
% \begin{tabular}{ccc}
% \toprule
% Method & Top-K & SST5 accuracy $\uparrow$   \\
% \midrule
% \midrule
% QLoRA FT(Baseline)             & 2    &  \textbf{53.6\%} \\
% \midrule
% QLoRA FT(Baseline)             & 1    &  50.8\% \\ 
% FT + Entropy loss, $\lambda=1$    & 1    & 45.5  \\
% FT + Entropy loss, $\lambda=0.1$    & 1    & 50.7\% \\
% + Annealing Top-K & 1 & 48.6\%    \\
% + Annealing Top-K (more steps)    & 1    &  51.4\% \\
% + Annealing Top-K + Entropy Loss    & 1    &  \textbf{51.8}\% \\
% \bottomrule
% \end{tabular}
% \caption{Comparison of extra strategies and their impact on SST5 accuracy.}
% \label{table:ft_sst5_ablation}
% \vspace{-15pt}
% \end{table}

\subsection{SEER-MoE: putting everything together}

We now explore whether combining both approaches yield additional gains.
We first sparsify the experts via the heavy-hitters counting and then finetune. Our goal with this is to understand how we can reduce FLOPs utilization without losing performance.

% \begin{table}[ht]
\begin{wrapfigure}{r}{0.7\textwidth}
\centering
\adjustbox{width=.7\textwidth}
{
\begin{tabular}{cccc}
\toprule
Counting / Pruning Strategy & Top-K FT Method & Top-K & SST5 acc $\uparrow$   \\
\midrule
\midrule
Activation / Global (25\%) &  QLoRA FT   &  2 & \textbf{49.0\%}  \\
\midrule
Activation / Global (25\%) &  QLoRA FT   &  1 & 47.5\%  \\
Soft / Global (25\%) &  QLoRA FT   &  1 & 46.7\%  \\
Activation / Global (25\%) &  \makecell{QLoRA FT \\ + Annealing Top-K  \\ + Entropy Loss}   &  1 & \textbf{48.0\%} \\
\bottomrule
\end{tabular}
}
\caption{Full stage approach results on SST5.}
\label{table:ft_sst5_ablation}
\vspace{-10pt}
% \end{table}
\end{wrapfigure}

Notably, our SEER-MoE approach, achieved a competitive accuracy of \textbf{48.0\%} while only activating a single expert. This highlights the potential of SEER-MoE to maintain high model performance even under significantly reduced computational overhead. Remarkably, the accuracy attained mirrors that of the two-expert configuration following only QLoRA fine-tuning, underscoring the effectiveness of our comprehensive strategy in reducing FLOPs without detriment to accuracy.

Furthermore, it is evident from the results that the activation-based pruning combined with a single-expert QLoRA fine-tuning confers a substantial accuracy gain over the soft pruning approach. This suggests that the more targeted activation-based pruning method combines effectively with our fine-tuning paradigm.

\section{Conclusion}

Our SEER-MoE framework effectively mitigates some of the computational inefficiencies of Mixture-of-Experts (MoE) models with minor compromise to performance. Through a two-stage process that includes expert sparsification and Top-K adaptation via fine-tuning, we have significantly cut down both FLOPs and memory usage for Mixtral 8x7b. Testing on SST5 and MMLU benchmarks shows that SEER-MoE achieves strong performance while reducing the number of active experts and parameters, making MoE models more viable across various applications and hardware constraints.

% \section*{Acknowledgement}

% We thank our mentor Kaylee Carissa Burns for being our mentor.

\bibliographystyle{acl_natbib}
\bibliography{references}

\clearpage
\appendix

\section{Parameters/FLOPs per Token for MoE Transformers}
\label{app:param_flops_analysis}

Table \ref{table:transformer-flops} describes the amount of parameters and FLOPs per token for a Transformer MoE model.

\begin{table}[ht]
\centering
\begin{tabular}{lcc}
\hline
\textbf{Operation} & \textbf{Parameters} & \textbf{FLOPs per Token} \\ \hline
Embed & \((n_{vocab} + n_{ctx}) d_{model}\) & \(4d_{model}\) \\
Attention: QKV & \(n_{layer}d_{model}3d_{attn}\) & \(2n_{layer}d_{model}3d_{attn}\) \\
Attention: Mask & — & \(2n_{layer}n_{ctx}d_{attn}\) \\
Attention: Project & \(n_{layer}d_{attn}d_{model}\) & \(2n_{layer}d_{attn}d_{embd}\) \\
\textbf{MoE Feedforward} & \(n_{experts} n_{layer}2d_{model}d_{ff}\) & \(2n_{topk} n_{layer}2d_{model}d_{ff}\) \\
\textbf{MoE Gating} & \(n_{experts}n_{layer}d_{model}\) & \(2n_{experts}n_{layer}d_{model}\) \\
De-embed & — & \(2d_{model}n_{vocab}\) \\ \hline
% \textbf{Total (Non-Embedding)} & \makecell{\(N = \\ 2d_{model}n_{layer} (2d_{attn} + n_{experts}d_{ff})\)} & \(C_{forward} = 2N + 2n_{layer}n_{ctx}d_{attn}\) \\ \hline
\end{tabular}
\caption{Parameter counts and compute (forward pass) estimates for a MoE Transformer model.  Nonlinearities, biases, and layer normalization are omitted.}
\label{table:transformer-flops}
\vspace{-15pt}
\end{table}

\section{Prompts}
\label{app:prompts}

In this section we describe the different prompts that we use for different tasks to query the desired answer.
We have experimented with different prompts and the ones presented below are chosen due to better performance or that match previously reported benchmarks.

\textbf{SST5} For Mixtral:\newline
\begin{verbatim}
[INST] You are a helpful assistant. Your task is of sentiment classification. 
Categorize the following text as either "very negative", "negative", 
 "neutral", "positive" or "very positive":
\{TEXT\} 
Only generate the label, without explanations:[/INST]
\end{verbatim}

\textbf{SST5} For OpenMoE:\newline
\begin{verbatim}
<<SYS>> You are a helpful assistant. Your task is of sentiment classification. <</SYS>>
<s>[INST] Categorize the following text in one of the following sentiments 
'very negative', 'negative', 'neutral', 'positive' or 'very positive':
\{TEXT\} [/INST]
\end{verbatim}

\textbf{MMLU} For Mixtral 8x7b:\newline
\begin{verbatim}
[INST] The following are multiple choice questions (with answers) about \{SUBJECT\}.

\{QUESTION\}

\{CHOICES\}

Only respond with the letter of the correct answer and no explanation.
Answer:
[/INST]
\end{verbatim}

\textbf{MMLU} For OpenMoE:\newline
\begin{verbatim}
<<SYS>> You are a helpful assistant. Your task is of multiple choice question answering
based on your knowledge. 
For subject \{SUBJECT\} 
Choose the best answer (A), (B), (C), or (D) 
 to the following question without explanation:<</SYS>>

<s>[INST] \{QUESTION\} from the following choices:

\{CHOICES\}

[/INST]
\end{verbatim}

\section{OpenMoE results}
\label{app:openmoe}

We also experimented with OpenMoE \cite{xue2024openmoe} models but we were not able to get reasonable results. 
We evaluate the  OpenMoE-8B-Chat model on MMLU and SST5 and the results can be seen in Table \ref{tab:initial_eval_openmoe}.

 \begin{table}[ht]
\centering
\begin{tabular}{|l|l|l|l|}
\hline
    \textbf{Model}                 & \textbf{Setup}     & \textbf{SST5 Acc.}  & \textbf{MMLU Acc.} \\ \hline
    OpenMoE-8B-Chat (1.1T+SFT)     & expert-topk=1      & 35.0\%                &  25.7\%   \\ \hline
    OpenMoE-8B-Chat (1.1T+SFT)     & expert-topk=2      & 35.6\%                &  26.5\%   \\ \hline
\end{tabular}
\caption{Evaluation on MMLU and SST5 for different for OpenMoE models with top-1 and top-2. }
\label{tab:initial_eval_openmoe}
\end{table}

Considering the random-guessing baseline for MMLU to be 25\% and for SST5 to be 20\%, we did not further pursue utilizing these models for additional experiments.

\end{document}